\documentclass[letterpaper, 10 pt, conference]{ieeeconf}  % Comment this line out if you need a4paper

\usepackage{cite}
\usepackage{subcaption} % loads the caption package
\usepackage{graphicx} % for images
\usepackage{float}
\usepackage{hyperref}  % Makes references clickable
\hypersetup{
    colorlinks=true,
    linkcolor=blue,
    citecolor=blue,
    urlcolor=blue
}
\usepackage{amsmath}
\IEEEoverridecommandlockouts                              % This command is only needed if 
\title{\LARGE \bf
Towards High Precision: An Adaptive Self-Supervised Learning Framework for Force-Based Verification
}

\author{Zebin Duan$^{1}$, Frederik Hagelskjær$^{1}$, Aljaz Kramberger$^{1}$, Juan Heredia$^{1}$, Norbert Krüger$^{1, 2}$% <-this % stops a space
    \thanks{
    $^{1}$All authors are with the Mærsk Mc-Kinney Møller Institute, University of Southern Denmark, 5230 Odense M, Denmark.
            {\tt\small \{zeb,frhag,alk,jehm,norbert\}@mmmi.sdu.dk}
    }%
    \thanks{$^{2}$The author is affiliated with the Danish Institute for Advanced Study (DIAS), University of Southern Denmark, 5230 Odense M, Denmark. {\tt\small norbert@mmmi.sdu.dk}}%
}
\begin{document}

\maketitle
\thispagestyle{empty}
\pagestyle{empty}

%%%%%%%%%%%%%%%%%%%%%%%%%%%%%%%%%%%%%%%%%%%%%%%%%%%%%%%%%%%%%%%%%%%%%%%%%%%%%%%%
\begin{abstract}
The automation of robotic tasks requires high precision and adaptability, particularly in force-based operations such as insertions. Traditional learning-based approaches either rely on static datasets, which limit their ability to generalize, or require frequent manual intervention to maintain good performances. As a result, ensuring long-term reliability without human supervision remains a significant challenge. To address this, we propose an adaptive self-supervised learning framework for insertion classification that continuously improves its precision over time. The framework operates in real-time, incrementally refining its classification decisions by integrating newly acquired force data. Unlike conventional methods, it does not rely on pre-collected datasets but instead evolves dynamically with each task execution. 
Through real-world experiments, we demonstrate how the system progressively reduces execution time while maintaining near-perfect precision as more samples are processed. This adaptability ensures long-term reliability in force-based robotic tasks while minimizing the need for manual intervention.
\end{abstract}

%%%%%%%%%%%%%%%%%%%%%%%%%%%%%%%%%%%%%%%%%%%%%%%%%%%%%%%%%%%%%%%%%%%%%%%%%%%%%%%%
\section{INTRODUCTION}

\begin{figure} [tb]
    \centering
    \includegraphics[width=0.70\linewidth]{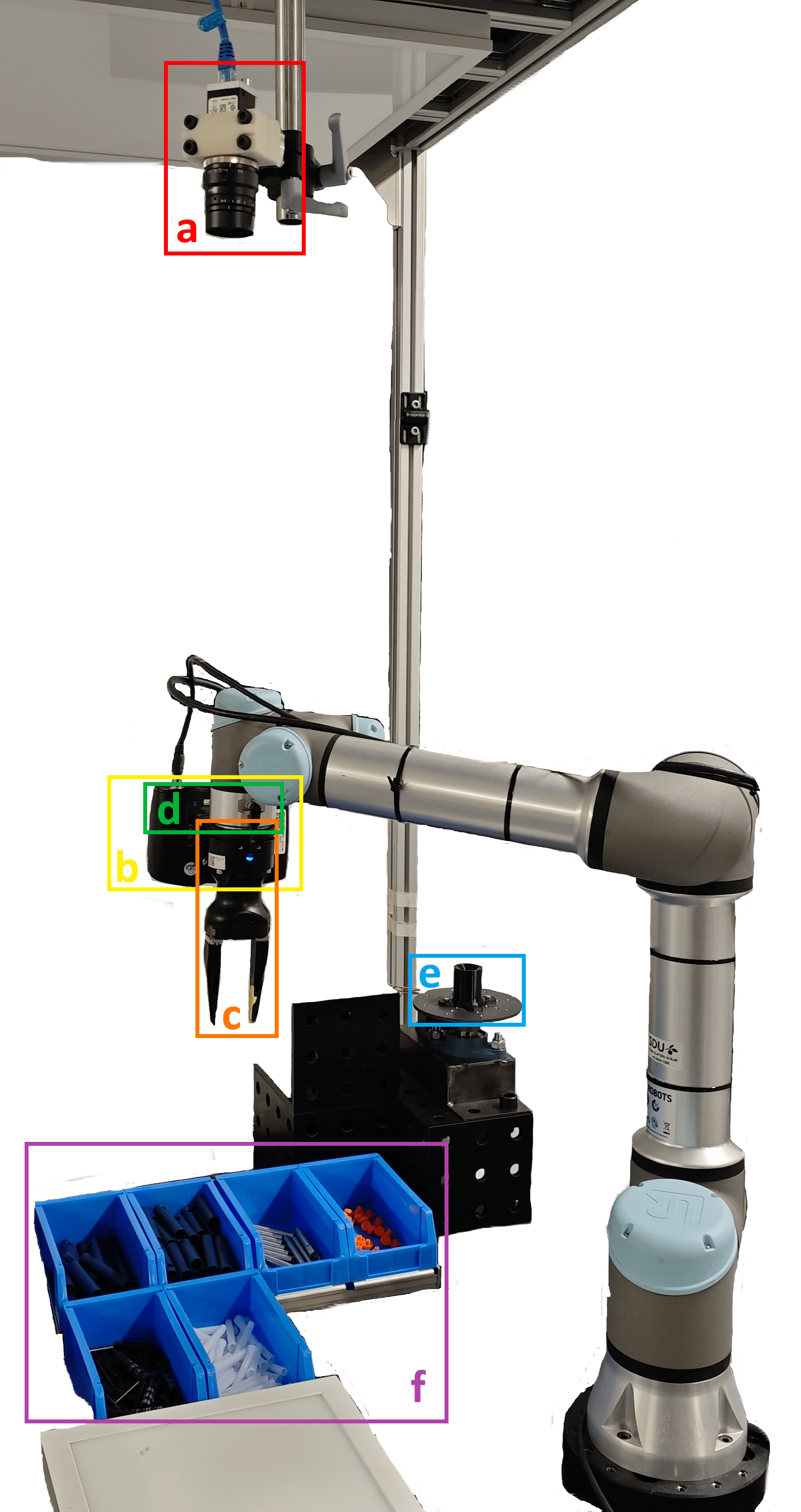}
    \caption{Work cell setup consisting of the following components: (a) 2D camera positioned above the work cell for auxiliary top-down view, (b) 3D camera mounted on the gripper for vision-based tasks, (c) Gripper with two fingers for object manipulation (d) Force-Torque sensor built in the robot end-effector (e) Fixture on top of a rotating table where the object will be inserted (f) Bins containing the test objects.}
    \label{fig:worksetup}
\end{figure}

Achieving high precision in force classification of robotic tasks presents a significant challenge. Traditional machine learning approaches require large-scale labeled datasets for supervised training, but manual labeling is often impractical for industrial applications. It is time-consuming, prone to inconsistencies when performed by non-experts, and unsuitable for high-mix low-volume production environments where rapid deployment is essential \cite{imagenet2015} \cite{degregoriosemiautomatic2020}. To overcome these limitations, self-supervised learning techniques have emerged \cite{jingselfsupervised2021}, enabling systems to autonomously generate labels without human intervention.

% Environment variations as in: the task where you're managing soft objects in contact, or task where you're doing insertion in a hole by feeling the force data, or admittance controller as you don't have the perfect 3D model of the object.
Force sensing is fundamental for robotic assembly \cite{leeAssemblyProcessMonitoring2019}, insertion \cite{peginholeshen2025}, and other contact-based interactions, as it allows robots to adapt to environment variations, detect anomalies \cite{anomalydetect2023} and improve task success rates. However, force measurements are highly sensitive to external factors such as temperature, friction and surface variations, making it challenging to develop accurate force-based classification models without real-world data. While simulations are widely used in robotics \cite{zhangbridging2024}, accurately modeling force interactions introduces significant complexity, reinforcing the necessity of real-world data for robust learning.

\begin{figure*}[t]
    \vspace{1.5mm}
    \centering
    \subcaptionbox{}
    {\includegraphics[width=0.19\textwidth]{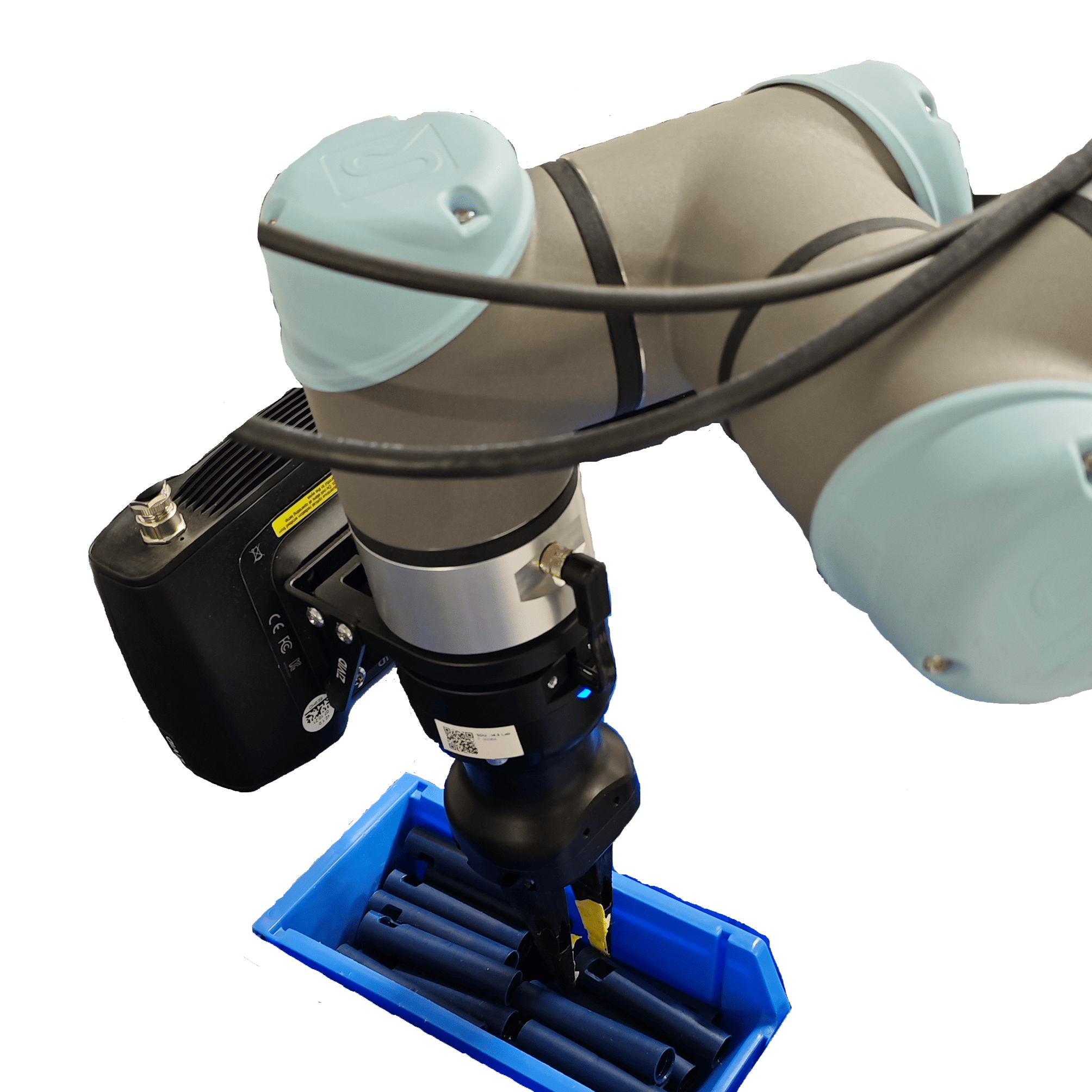}}
    \hfill
    \subcaptionbox{}
    {\includegraphics[width=0.19\textwidth]{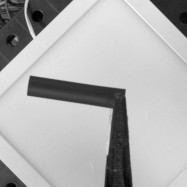}}
    \hfill
    \subcaptionbox{}
    {\includegraphics[width=0.19\textwidth]{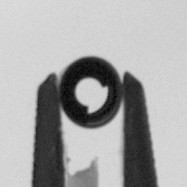}}
    \hfill
    \subcaptionbox{}
    {\includegraphics[width=0.19\textwidth]{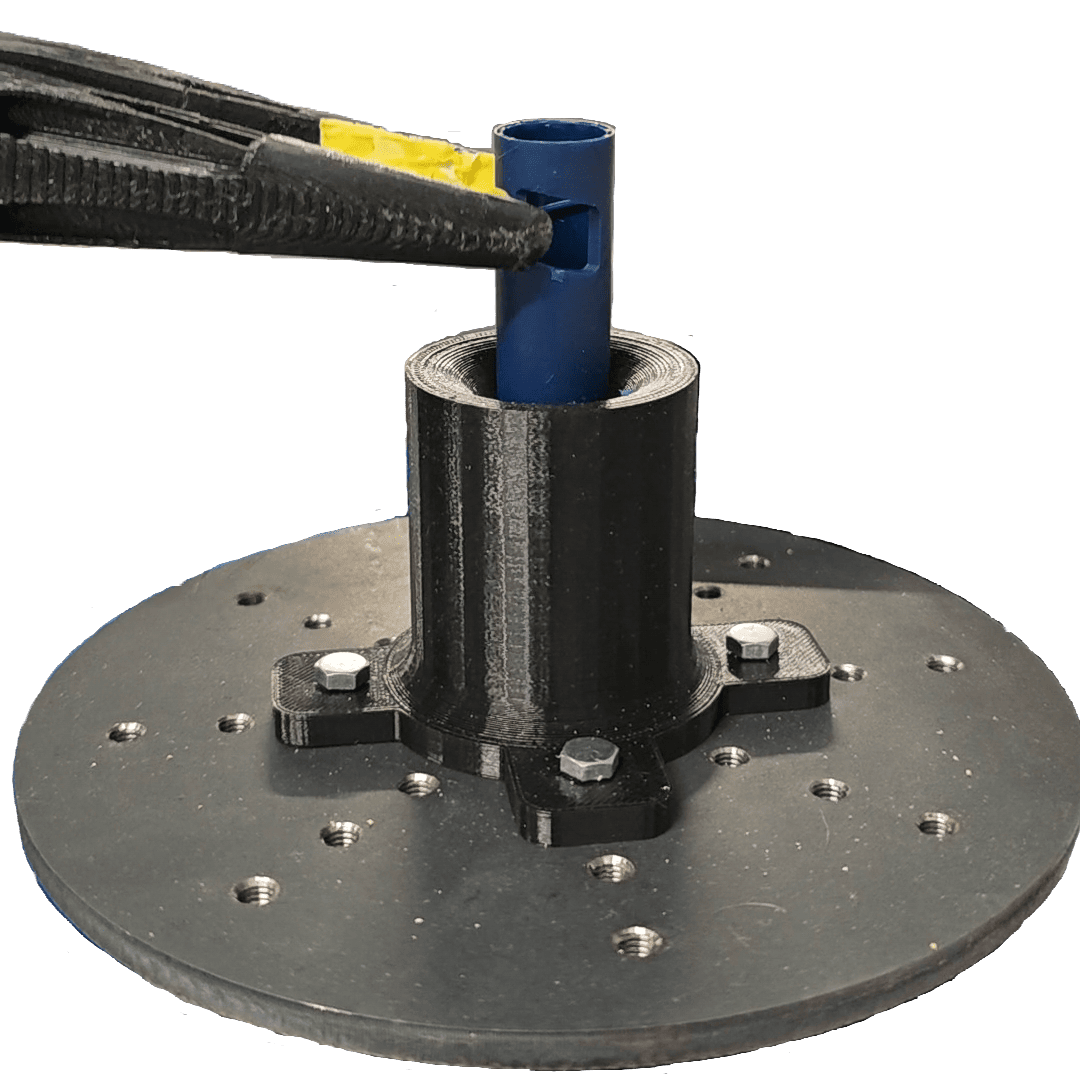}}
    \hfill
    \subcaptionbox{}
    {\includegraphics[width=0.19\textwidth]{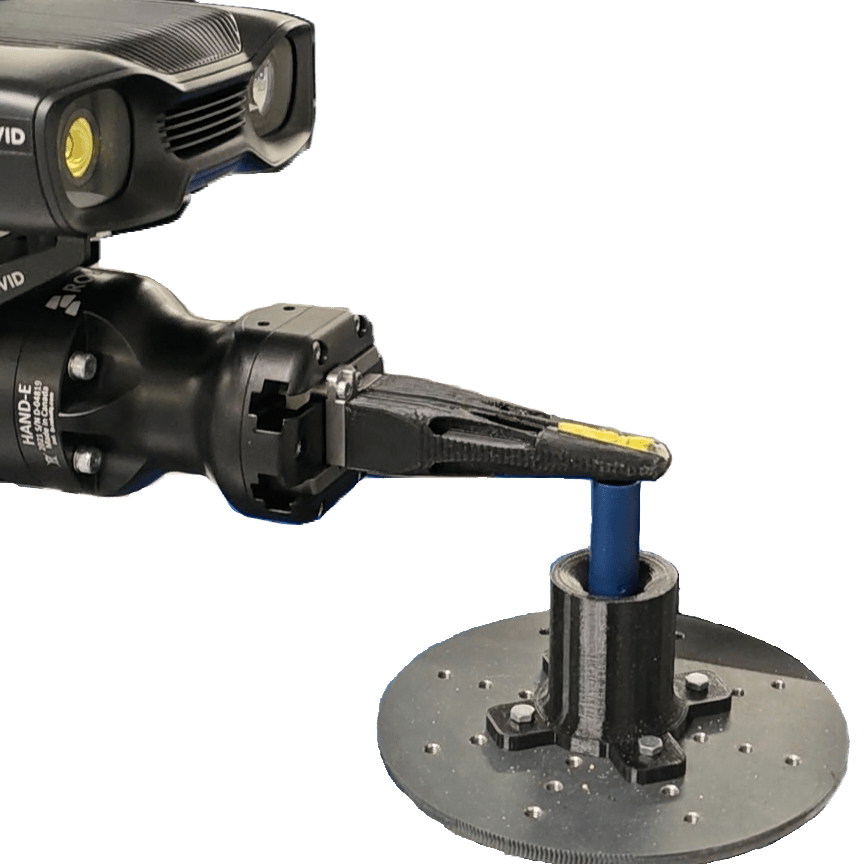}}
    \caption{The pipeline of our system: (a) Object pose estimation and bin picking of the object. (b) In-hand pose estimation (c) Revolution estimation from top-down view (d) Insertion of the object into the fixture. (e) Height verification.}
    \label{fig:pipeline}
\end{figure*}

To achieve high precision in classification tasks without relying on tedious manual labeling and simulations, we introduce an adaptive self-supervised learning framework that leverages force data. The framework incrementally improves insertion classification performances by continuously updating its model during execution. This approach eliminates the need for offline retraining, pre-collected datasets or manual annotations. 
% The system operates using the configuration described in Fig. \ref{fig:worksetup} and the workflow pipeline of Fig. \ref{fig:pipeline}. 
The system is tested using a bin picking work cell described in Fig. \ref{fig:worksetup} and the workflow pipeline of Fig. \ref{fig:pipeline}.
Unlike static models, our approach starts with minimal prior knowledge and refines its decision-making as new force data become available, progressively reducing misclassifications.

Our work demonstrates how force-torque sensors can be utilized to develop self-improving robotic workflows without additional infrastructure or costly validation steps. By applying an adaptive classification rule (Sec. \ref{subsec:votingrule}), the proposed framework allows robots to autonomously perform insertion tasks while enhancing performance over time.

Specifically, our contributions include:

\begin{itemize} 
    \item A method for dynamically improving insertion classification by continuously refining force-based decision-making.
    \item A self-supervised learning approach that enables long-term precision improvement in force classification tasks.
    \item Real-world experiments validating the system’s ability to progressively enhance precision with each execution.
\end{itemize}

The remainder of this paper details related works Sec. \ref{sec:relatedworks}, the methodology Sec. \ref{sec:methodology}, the experimental evaluation Sec. \ref{sec:experimentalevaluation}, and conclusions Sec. \ref{sec:conclusion}.

%%%%%%%%%%%%%%%%%%%%%%%%%%%%%%%%%%%%%%%%%%%%%%%%%%%%%%%%%%%%%%%%%%%%%%%%%%%%%%%%%%%%%

\section{RELATED WORK}
\label{sec:relatedworks}

Robotic insertion and assembly tasks pose persistent challenges due to the challenges of achieving both high precision and robust performance in uncertain environments. While progress in grasping has improved object acquisition, precise insertion remains significantly more sensitive to small errors, requiring robust feedback mechanisms and adaptive learning strategies.

%Robotic assembly and insertion tasks have gained significant attention in modern industry due to the challenges of achieving both high precision and robust performance in uncertain environments. In \cite{hagelskjaerGoodGraspsOnly2025}, the authors introduce a novel pipeline that leverages a self-supervised fine-tuning data engine for pose estimation, using grasp poses as a verification mechanism. This approach continuously refines the robot’s grasping strategy, significantly improving pose estimation accuracy and enhancing insertion performance. However, despite achieving a high grasp success rate, where only around five percent of detected objects fail to be grasped, this method does not guarantee a correct or error-free assembly, as successful grasping does not necessarily ensure accurate alignment or placement. This limitation highlights the necessity of additional verification stages, such as a dedicated testbed, to rigorously assess grasp quality.

\textit{Force based monitoring and verification} have been widely explored for insertion tasks. In industrial contexts, force-torque sensing is commonly used to detect deviations or insertion failures, as shown by Lee et al. \cite{leeAssemblyProcessMonitoring2019}, who monitor characteristic force profiles during component alignment. In surgical robotics, similar principles apply; for example, Pile et al. \cite{pileRobotassistedPerceptionAugmentation2017} use force feedback to detect critical failure points during cochlear electrode insertion. Pitchandi et al. \cite{pitchandiInsertionForceAnalysis2017} focus specifically on peg-in-hole tasks with compliant fixtures, identifying force signatures that differentiate successful from failed insertions. These works emphasize the importance of multimodal feedback, integrating force and tactile sensing, to enhance fault prediction in robotic solutions.

%Recent research has focused on leveraging force-torque data and tactile feedback to detect failures and enhance task execution in assembly and insertion tasks. For instance, in the domain of surgical robotics, force data combined with predictive models has been used to detect real time insertion failures due to insufficient support of the distal tip of the electrode array on the cochlea\cite{pileRobotassistedPerceptionAugmentation2017}. Similarly, force-based monitoring techniques have been employed in industrial applications to identify deviations during assembly processes \cite{leeAssemblyProcessMonitoring2019}. Furthermore, the study presented in \cite{pitchandiInsertionForceAnalysis2017} focuses specifically on the peg-in-hole assembly task under compliant support conditions. The authors investigate how the insertion force profile is affected by the compliant nature of the assembly fixture. Through detailed force analysis, they identify critical thresholds and patterns that differentiate between successful and failed insertions. Their findings emphasize the importance of multimodal feedback, integrating force and tactile sensing, to enhance fault prediction in robotic solutions. This approach has broad applications, ranging from medical robotics and precision assembly to large-scale industrial automation.

\textit{Learning-based approaches} have sought to generalize such force-torque driven verification. Zhou et al. \cite{zhouRoboticStackingIrregular2023} train supervised models on force-torque data to classify stacking outcomes, while Papavasileiou et al. \cite{papavasileiouQualityControlManufacturing2025} apply decision trees for contact-based quality control. However, such approaches often require extensive labeled datasets and may not adapt well to unseen configurations.

To mitigate these limitations, \textit{Self-supervised learning} has gained significant attention in robotic automation due to its ability to reduce reliance on manual labeling, allowing systems to continuously improve by learning directly from their own experiences. Berscheid et al. \cite{berscheidImprovingDataEfficiency2019} emphasize how this approach minimizes human supervision, particularly in robotic grasping tasks, where it enhances pose estimation and improves success rates over time. In robotic perception, self-supervised learning has been applied to visual modalities \cite{robertssonImplementationIndustrialRobot2006}, but only limited attention has been paid to self-supervised learning on force data. Moreover, \cite{hsuehCompactCompliantRobot2025} proposed an online learning framework for robotic grasp adaptation, though it lacked real-time force-based verification.

%In force-based robotic verification, several studies have explored supervised learning techniques, such as \cite{zhouRoboticStackingIrregular2023}, which used Convolutional Neural Networks (CNNs) trained on labeled force data. While this method achieved high accuracy, it required extensive manual annotation. Similarly, \cite{papavasileiouQualityControlManufacturing2025} employed a decision tree classifier for contact-based object recognition, but its performance was heavily dependent on predefined datasets. To overcome these challenges, self-supervised learning has been investigated in robotic perception with approaches like \cite{robertssonImplementationIndustrialRobot2006} using contrastive learning to generate pseudo-labels, primarily for visual data rather than force sensing. Moreover, \cite{hsuehCompactCompliantRobot2025} proposed an online learning framework for robotic grasp adaptation, though it lacked real-time force-based verification. Our work builds upon these studies by applying self-supervised learning to force data in insertion tasks, enabling the system to refine its predictions autonomously and continuously enhance its performance in dynamic environments. 

Although previous work has explored \textit{force sensing} and \textit{self-supervised learning} independently, few studies have combined these approaches for real-time robotic verification. Our work builds upon these studies by applying self-supervised learning to force data in insertion tasks, enabling the system to refine its predictions autonomously and continuously enhance its performance dynamically to new instances.

%Our proposed framework integrates force-based self-supervised learning, enabling automatic instance labeling while continuously improving classification precision. Unlike existing methods, our approach minimizes the need for manual labelling and adapts dynamically to new instances.

\section{METHODOLOGY}
\label{sec:methodology}

\subsection{System overview}
The workflow pipeline is shown in Fig. \ref{fig:pipeline}. The system is based on the improved bin-picking work cell of \cite{hagelskjaer2024off} introduced in \cite{hagelskjaerGoodGraspsOnly2025}. 
In the first stage, the system captures a 3D point cloud of the scene, estimates the poses of objects inside the bin, and computes grasp poses as seen in Fig. \ref{fig:pipeline}(a). A grasp attempt is then performed, placing the object in the robot's fingers. This is followed by in-hand pose estimation as shown in Fig. \ref{fig:pipeline}(b), which corrects any misalignment that may have occurred during grasping. Finally, the object is placed into a fixture based on the found pose estimate. 

For the object addressed in this paper, the in-hand pose estimation is inadequate. To insert the object correctly into the fixture, the full 6 Degrees of Freedom pose estimate should be known. However, the correct orientation of the object is uncertain given the view of the in-hand pose estimation, see \ref{fig:pipeline}(b). To determine the revolution of the object, a second in-hand pose estimation is performed, where the object is rotated to present a view of the internal structure, see Fig. \ref{fig:pipeline}(c). This rotation is based on the initial in-hand pose estimate. The second in-hand pose estimate is performed using the same template matching as the initial in-hand pose estimate \cite{hagelskjaer2022hand}. The object is then placed into the fixture, where the fixture has been placed on a rotating table allowing the object to be placed in the correct orientation as seen in Fig. \ref{fig:pipeline}(d). Lastly, the system performs a height verification to classify the insertion as successful or unsuccessful Fig. \ref{fig:pipeline}(e). 

Unlike \cite{hagelskjaerGoodGraspsOnly2025}, during the insertion step in Fig. \ref{fig:pipeline}(d), we continuously record force data from the force-torque sensor, enabling immediate classification of the insertion upon completion using the classification algorithm based on $k$-NN and $l$-Value voting rule described in Sec. \ref{subsec:votingrule}. The time-cost height verification method in Fig. \ref{fig:pipeline}(e) is only triggered when the model's prediction is uncertain, reducing the need for this additional verification step and optimizing overall processing time.

\begin{figure} [htb]
    \vspace{1.5mm}
    \centering
    \includegraphics[width=0.425\textwidth]{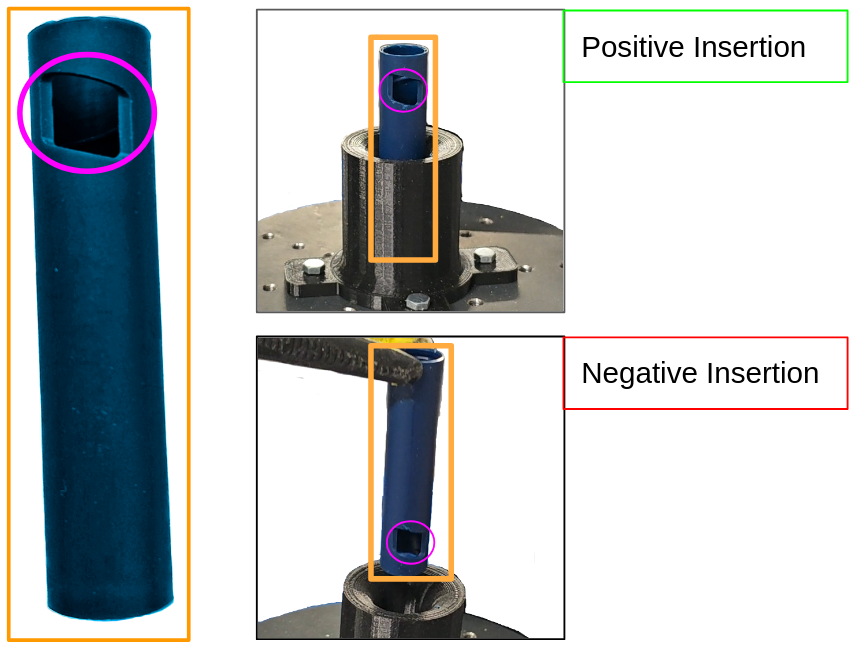}
    \caption{Test object overview, part of an injection device}
    \label{fig:testobjdescr}
\end{figure}

The test object used in our experiments is illustrated in Fig. \ref{fig:testobjdescr}. It has a single lateral hole, and the insertion success is determined based on the object's post-insertion height. Specifically, when the lateral hole is positioned on the top side after insertion, the object's height is lower compared to when the hole is on the bottom side. In this case, the insertion is classified as positive; otherwise, it is considered negative. The hole’s position can also be inferred from a top-down view as we can see from Fig. \ref{fig:pipeline}(c). However, this approach only determines the lateral hole's correct position without indicating whether it is on the top or bottom of the cylindrical object.
In our experiments, we will only consider the forces in the z-axis as the main movement is along the z direction.

\subsection{Data Collection Characteristics}
\label{subsec:datatcollection}
\begin{figure}[tb]
    \centering
    \vspace{1.5mm}
    \includegraphics[width=0.48\textwidth]{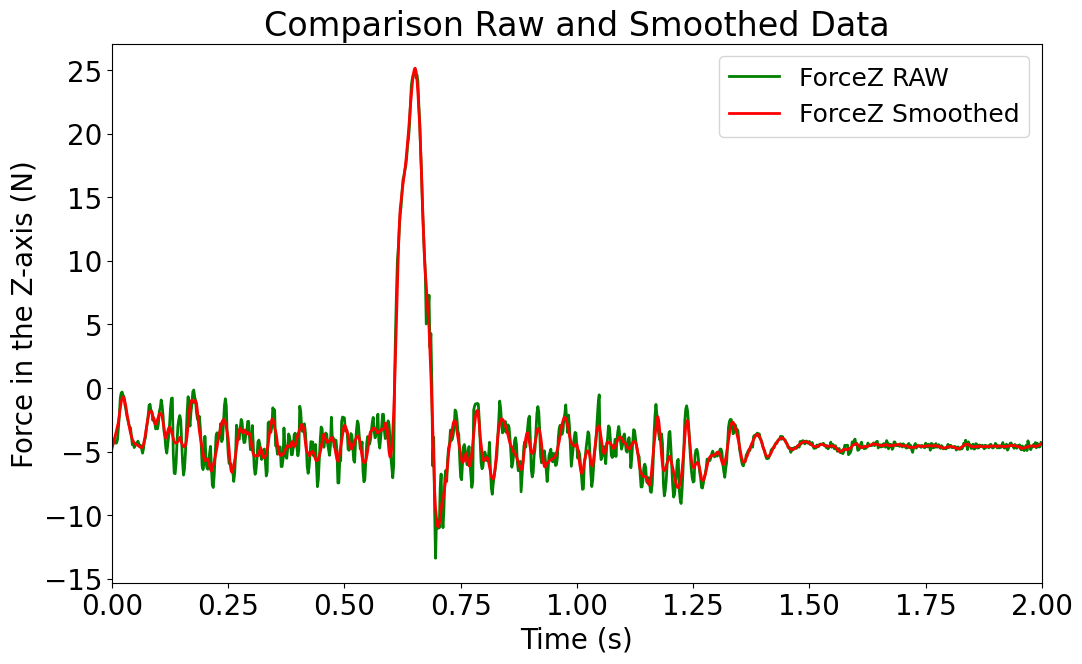}
    \caption{Denoising effect of the Savitzky-Golay filter on a negative insertion force profile.}
    \label{fig:savgolfilter}
\end{figure}

Force data was recorded at 500 Hz over a 2-second interval, resulting in 1000-dimensional feature vectors. To mitigate sensor noise, we applied a Savitzky-Golay filter \cite{savgov2011} with $window\_size = 15$ and $polynomial\_order = 2$, which effectively preserves force peaks while attenuating high-frequency noise as we can see from Fig. \ref{fig:savgolfilter}. 
Finally, to reduce the dimensionality of the feature vectors, we employed a sliding window mean down-sampling approach.

\subsection{k-NN and $l$-Value voting rule}
\label{subsec:votingrule}
With the collected time-series force data we implement the k-Nearest Neighbors ($k$-NN) algorithm from Scikit-Learn \cite{scikitlearn2011} for this task. Upon receiving a new data point, the $k$-NN algorithm computes its distance to all points in the training dataset using a distance metric, typically the euclidean distance. The algorithm then identifies the $k$ nearest points in the feature space, where $k$ is a user-defined parameter. The class label most frequent among these $k$ neighbors is assigned to the new data point. The selection of $k$ and the distance metric plays a critical role in the performance of the algorithm. A smaller $k$ value may make the model more sensitive to noise, while a larger $k$ value can lead to a smoother decision boundary.

While $k$-NN classification traditionally relies on a majority voting mechanism, we propose a modified decision rule that incorporates a minimum agreement threshold, denoted as the $l$-Value. Specifically, the $l$-Value requires a predefined percentage of the $k$ neighbors to agree on the same class label for a prediction to be accepted. For example, a $l$-Value of 90\% requires that at least 90\% of the nearest neighbors must agree on the classification outcome. If this consensus threshold is not met, the system abstains from making a definitive prediction and instead defaults to the verification process outlined in Fig. \ref{fig:pipeline}(e) for final validation.

Mathematically, the modified decision rule can be expressed as:

\begin{equation}
    \frac{N_c}{k } * 100 \ge l \%
\end{equation}

where $N_c$ is the number of neighbors that agree on the class label, $k$ is the total number of nearest neighbors, and $l$ is the $l$-Value (expressed as a percentage). 

For instance, with $k=11$, at least 10 neighbors must agree on the classification outcome for $l$-Value of 90\%, that is $N_c  >= 9.9$. 

\begin{figure}[thbp]
    \centering
    \vspace{1.5mm}
    \subcaptionbox{}{\includegraphics[width=0.48\textwidth]{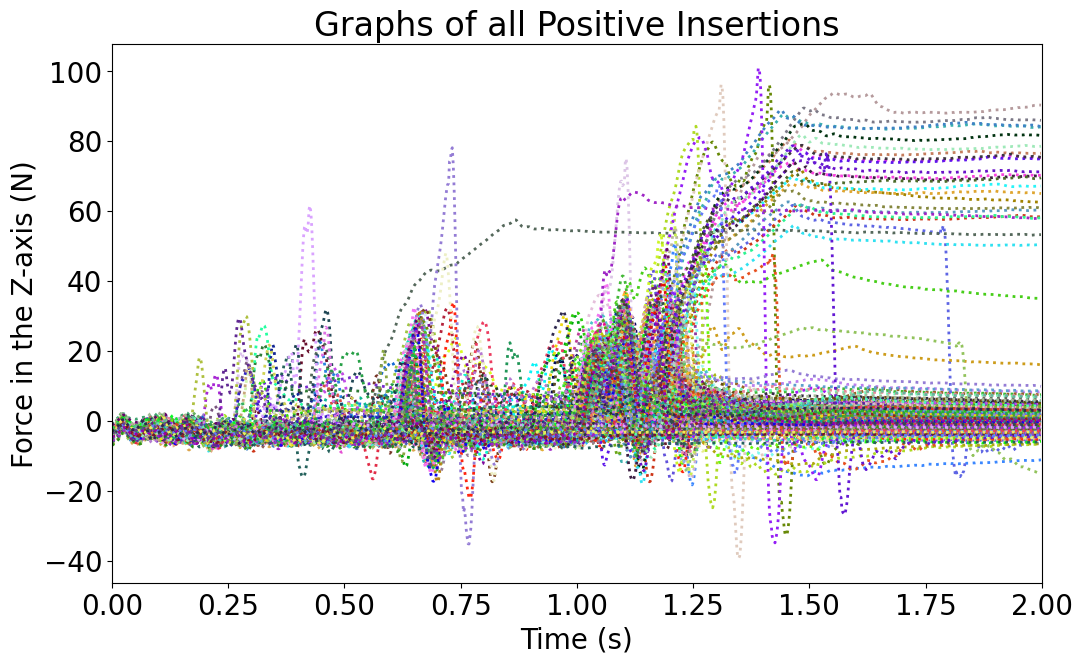}}%
    \\
    \subcaptionbox{}{\includegraphics[width=0.48\textwidth]{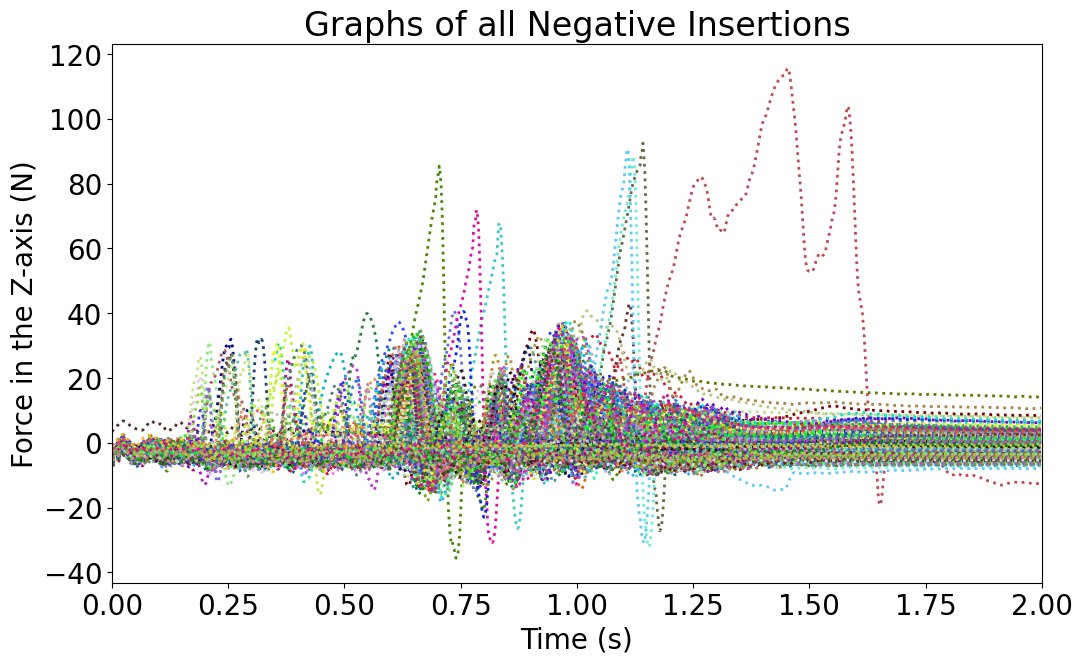}}%
    \caption{All force data profiles along the z-axis over time, divided by their class: (a) Positive insertions and (b) Negative insertions.}
    \label{fig:Dataset}
\end{figure}

In this study, we set a fixed $k=11$ to maintain consistency and simplicity in the classification process, avoiding the added complexity of optimizing $k$. Additionally, given the high-dimensional nature of our feature vector, we opted for cosine distance instead of the euclidean distance. Cosine distance is well-suited for high-dimensional data and performs effectively with sparse data, as observed in Fig. \ref{fig:Dataset} and from the results in Tab. \ref{tab:gridsearchresults}.

\section{EXPERIMENTAL EVALUATION}
\label{sec:experimentalevaluation}

\subsection{Datasets}
\label{subsec:datasets}
We collected a total of 704 insertion trials shown in Fig. \ref{fig:Dataset}, comprising 297 positive insertions and 407 negative insertions. Of these, 604 trials were used for training and validation, while the remaining 100 were set aside for testing.

\subsection{Evaluation metrics}
We evaluated our model using precision and recall Eq. \ref{lbl:precrec}, which are defined in terms of True Positives (TP), True Negatives (TN), False Positives (FP) and False Negatives (FN), and described in Tab. \ref{tab:confusion_matrix}:

\begin{equation}
\begin{aligned}
    %Accuracy &= \frac{TP + TN}{TP + TN + FP + FN}\\
    Precision &= \frac{TP}{TP + FP}\\
    Recall &= \frac{TP}{TP + FN}\\
    %F_{1} Score &= 2 \times \frac{Precision \times Recall}{Precision + Recall}
\end{aligned}
\label{lbl:precrec}
\end{equation}

\begin{table}[h]
    \centering
    \vspace{1.8mm}
    \caption{Confusion Matrix for Insertion Classification}
    \renewcommand{\arraystretch}{2.0}
    \resizebox{\columnwidth}{!}{
    \begin{tabular}{|p{1.5cm}|p{2.8cm}|p{2.8cm}|} 
        \hline
         & \textbf{Actual Positive} & \textbf{Actual Negative} \\
        \hline
        \textbf{Predicted Positive} & 
        \textbf{TP}: Correctly classified positive insertion & 
        \textbf{FP}: Incorrectly classified positive insertion, actual insertion is negative \\
        \hline
        \textbf{Predicted Negative} & 
        \textbf{FN}: Incorrectly classified negative insertion, actual insertion is positive & 
        \textbf{TN}: Correctly classified negative insertion \\
        \hline
    \end{tabular}
    }
    \label{tab:confusion_matrix}
\end{table}

% Where???
% As discussed in Sec. \ref{subsec:votingrule},
Our primary objective is to maximize precision, as this reduces the occurrence of false positives, which could result in assembly defects or safety hazards, potentially affecting subsequent stages of the insertion process. 
% In our specific case, false positives can lead to \textcolor{red}{malfunctioning insulin insertion devices}, resulting in improper insertion and complications during injection. These issues can cause significant health problems for patients. 
In our specific real case, false positives can lead to malfunctioning injection devices. 
In contrast, false negatives pose minimal risk, as they only incur a slight time cost for removing the object from the fixture after the insertion. By prioritizing precision, only high-confidence insertions are classified as successful, effectively minimizing the propagation of errors in downstream operations.

\subsection{Performance Evaluation Metrics}
\begin{figure*}[thpb]
    
    \centering
    
    \hfill
    \vspace{1.8mm}
    \hfill
    \\
    % \captionsetup[subfigure]{labelformat=simple, labelsep=period}
    \subcaptionbox{}
    {\includegraphics[width=0.315\textwidth]{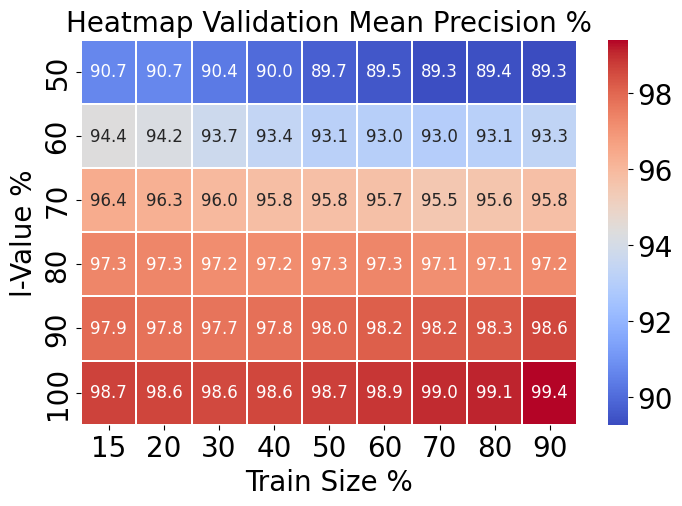}}
    \hfill
    \subcaptionbox{}
    {\includegraphics[width=0.315\textwidth]{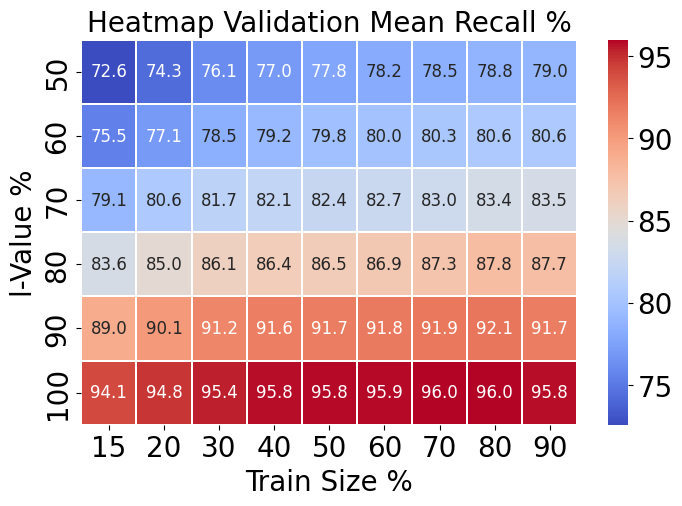}}
    \hfill % <-- Line break
    \subcaptionbox{}{\includegraphics[width=0.315\textwidth]{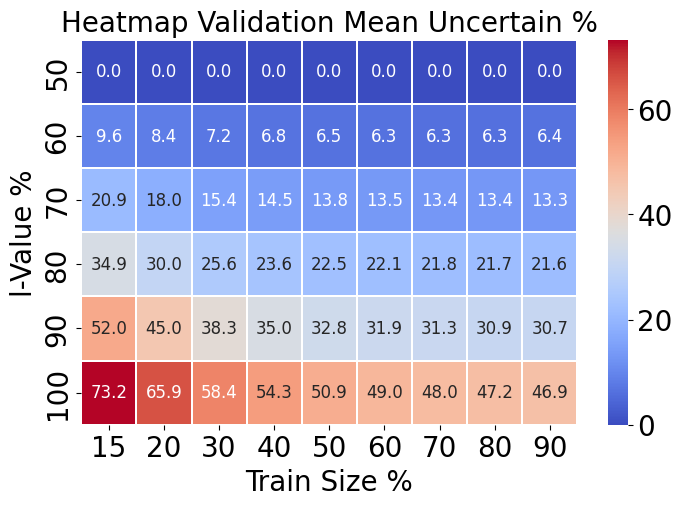}}%
    \\ % <-- Seperation
    \subcaptionbox{}
    {\includegraphics[width=0.315\textwidth]{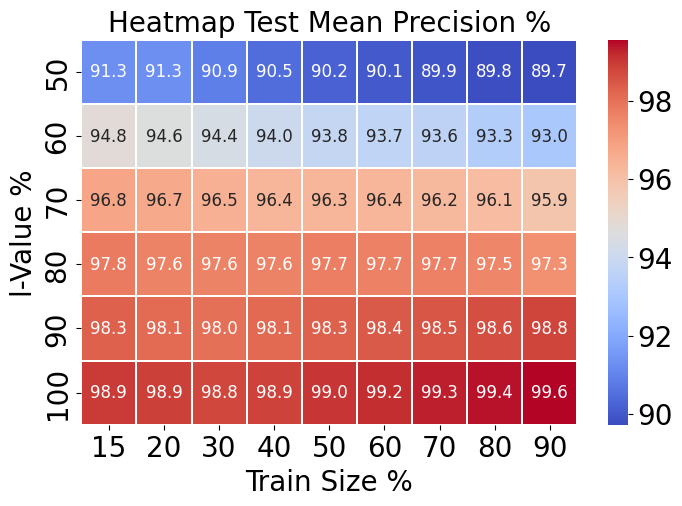}}%
    \hfill
    \subcaptionbox{}
    {\includegraphics[width=0.315\textwidth]{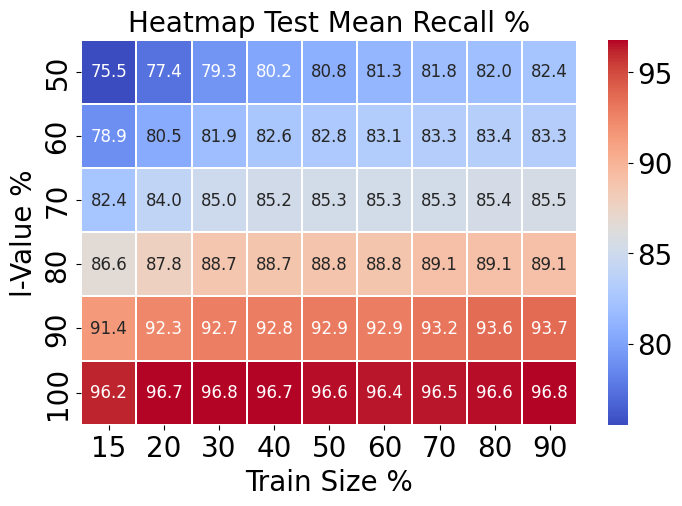}}%
    \hfill
    \subcaptionbox{}
    {\includegraphics[width=0.315\textwidth]{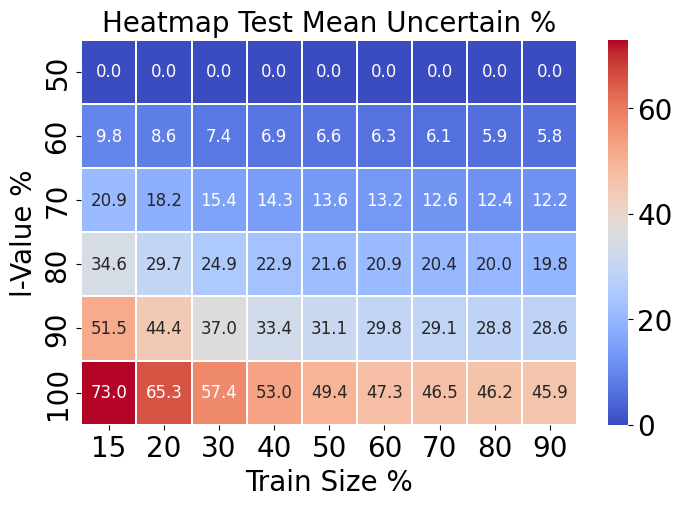}}%
    \caption{Heatmap results of evaluation metrics and uncertain percentages for Validation and Test dataset with respect to different training size and $l$-Value. In order: Validation Set (a) Precision, (b) Recall and (c) Uncertain; Test Set (d) Precision, (e) Recall and (f) Uncertain.}
    \label{fig:heatmaps}
\end{figure*}

We performed experiments by varying the training set size, random seed values, and classification $l$-Value while preserving the training, validation, and test ratio outlined in Sec. \ref{subsec:datasets}. The results were then aggregated using the mean function and are presented in Fig. \ref{fig:heatmaps}.

The increase of training size leads to an improvement in model performance. However, the most notable finding is the significant performance gain achieved by modifying the voting rule, as described in Sec. \ref{subsec:votingrule}. This improvement is evident when comparing the first and last rows of the heatmaps in Fig. \ref{fig:heatmaps}(a)(b)(d)(e). As expected, this enhancement comes at the cost of a higher number of unclassified instances, as illustrated in Fig. \ref{fig:heatmaps}(c)(f). Specifically, according to the standard voting rule ($l-Value >= 50\%$), every instance is assigned a label, and so the model is never uncertain about the classification. In contrast ($l-Value >= 100\%$), when the training size is only 15\% of its original size, the model struggles to classify most instances, that is around 73\% of either the validation or test set. However, as the training set increases, the number of uncertain classifications decreases, thereby reducing the reliance on the time-cost height verification method as it is shown in Fig. \ref{fig:pipeline}(e).

\subsection{Online self-supervised performance}
% We evaluated our framework in a prototype online environment, where the training dataset was sequentially reused as new force data for classification. The framework was executed 30 times, with the dataset shuffled in each run to ensure variation in the initial base-growing dataset. To enhance reliability, final results were averaged across all runs. Each force signal was classified according to the method outlined in Sec. \ref{subsec:votingrule}.
We evaluated our framework in a prototype online environment. The framework was executed 30 times, with the dataset shuffled in each run to ensure variation in the initial base-growing dataset and as if encountering it for the first time from each run. To enhance reliability, final results were averaged across all runs. Each force signal was classified according to the method outlined in Sec. \ref{subsec:votingrule}.

Since the $k$-NN model initially lacks ground truth data, the system employs the height verification method Fig. \ref{fig:pipeline}(e) for the first instances. To maintain balance, at least half of this initial dataset consists of positive insertions. These verified samples are then incorporated into the base-growing dataset, enabling subsequent classifications to rely on an expanding reference set.

When the model encounters uncertainty in classification, the height verification method is triggered, and the newly labeled data is added to the base dataset. The model undergoes periodic retraining after a predefined number of classified instances (every 20 in our case), 
% BEFORE
% with the optimal $k$ value recomputed at each retraining step between a limited number of choices (odd numbers between 5 and 29). 
with fixed $k = 11$. 
After processing all instances, predicted labels were compared against ground truth for evaluation.

\begin{figure}[thbp]
    \vspace{1.5mm}
    \centering
    \subcaptionbox{}{\includegraphics[width=0.45\textwidth]{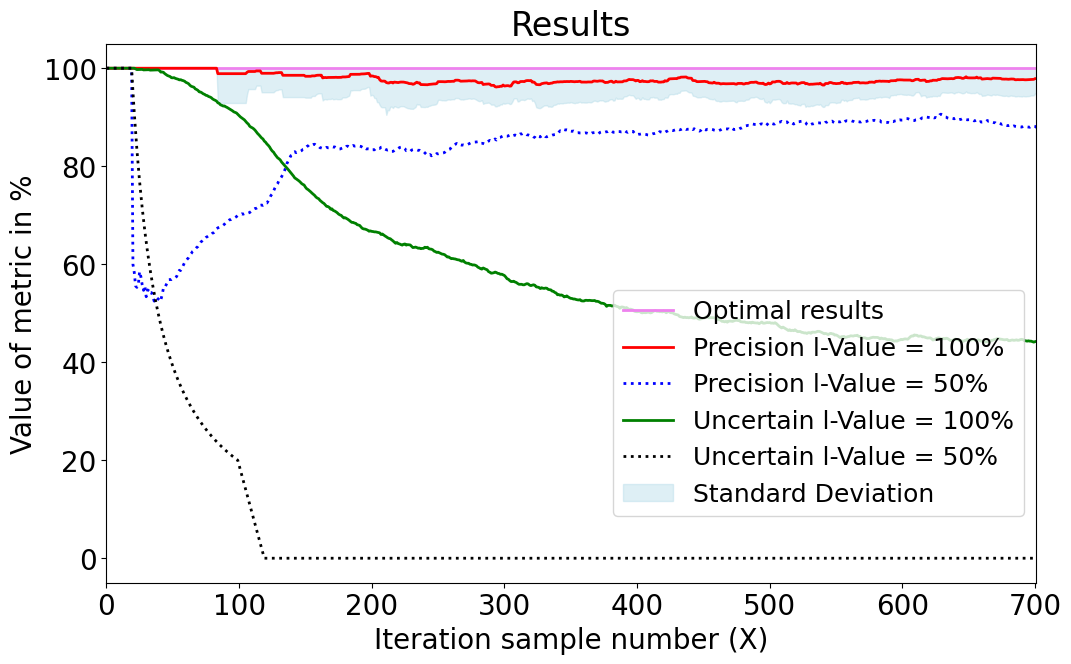}}
    \\%
    \subcaptionbox{}{\includegraphics[width=0.45\textwidth]{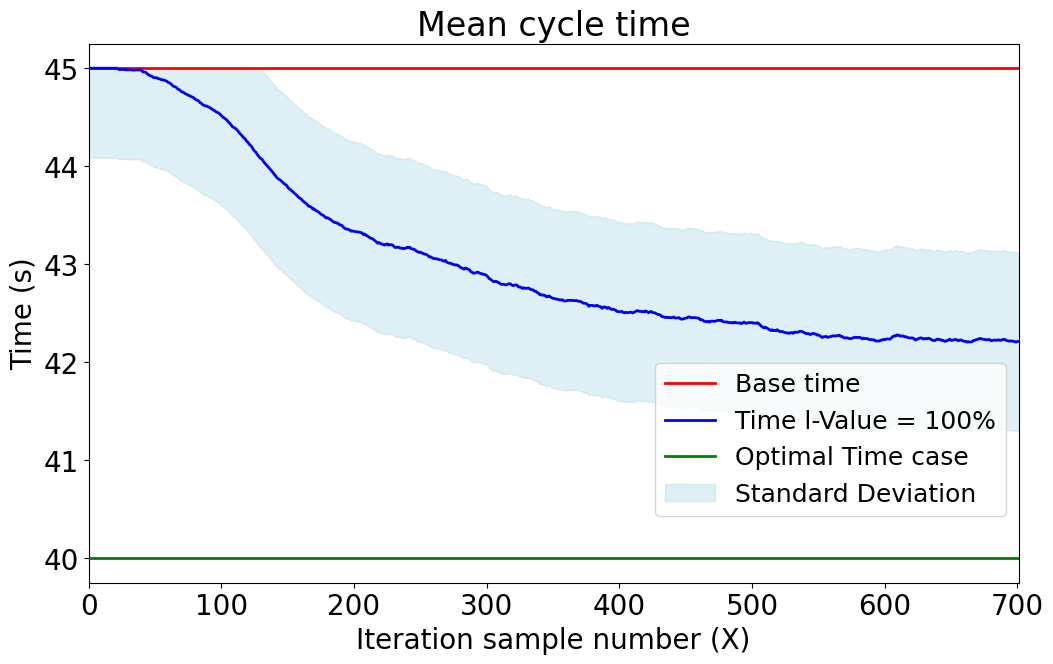}}
    \caption{Mean results of Online self-supervised framework performances over a sliding windows of 100 samples: (a) Results of Precision and Uncertain samples with different $l$-Value and (b) Mean cycle time execution.}
    \label{fig:resultsTraining}
\end{figure}

Fig. \ref{fig:resultsTraining}(a) illustrates model performance over a sliding window of 100 samples. At each iteration, precision and uncertainty percentages were computed based on the preceding 100 samples. These results indicate that, given the same base training dataset, precision is higher when $l$-Value = 100\% compared to when predicting with $l$-Value = 50\%, although at the cost of increasing reliance on the height verification method. However, as the base-growing dataset expands, the number of uncertain classifications progressively decreases.

% FIRST SUBMISSION TABLE
% \begin{table}[h]
% \caption{Mean Results of the simulated Online Self-Supervised Learning framework.}
% \begin{center}
%     \begin{tabular}{|c|c|c|}
%     \hline
%      & $l$-Value = 100\% & $l$-Value = 50\% \\
%     \hline
%     Samples & \multicolumn{2}{c|}{704}\\ 
%     \hline
%     Growing Base Dataset size & \multicolumn{2}{c|}{390.2}\\ 
%     \hline
%     Count Height Verification  & 390.2 & 704\\ 
%     \hline
%     Precision & 96.53\% & 84.59\%\\
%     \hline
%     Recall & 93.05\% & 82.57\%\\
%     \hline
%     Count of TP & 113.8 & 235.46 \\
%     \hline
%     Count of FP & 4.16 & 43.0 \\
%     \hline
%     Count of TN & 187.1 & 355.83 \\
%     \hline
%     Count of FN & 8.73 & 49.7 \\
%     \hline
%     \end{tabular}
% \end{center}
% \label{tab:onlineresults}
% \end{table}
\begin{table}[h]
\caption{Mean Results of the simulated Online Self-Supervised Learning framework.}
\begin{center}
    \begin{tabular}{|c|c|c|}
    \hline
     & $l$-Value = 100\% & $l$-Value = 50\% \\
    \hline
    Samples & \multicolumn{2}{c|}{704}\\ 
    \hline
    Growing Base Dataset size & \multicolumn{2}{c|}{401.7}\\ 
    \hline
    Count Height Verification  & 401.7 & 704\\ 
    \hline
    Precision & 97.72\% & 83.60\%\\
    \hline
    Recall & 94.50\% & 83.24\%\\
    \hline
    Count of TP & 110.2 & 237.4 \\
    \hline
    Count of FP & 2.6 & 46.8 \\
    \hline
    Count of TN & 182.8 & 351.8 \\
    \hline
    Count of FN & 6.4 & 47.8 \\
    \hline
    \end{tabular}
\end{center}
\label{tab:onlineresults}
\end{table}

Compared to previous work \cite{hagelskjaerGoodGraspsOnly2025}, which required the time-intensive height verification step for every instance, our approach significantly reduces its usage.

Assuming that the entire pipeline in Fig. \ref{fig:pipeline} takes approximately 45 seconds per iteration, with the height verification step contributing 5 seconds, we obtain the results shown in Fig. \ref{fig:resultsTraining}(b), computed over a sliding window of 100 samples. 

% BEFORE
% The overall classification results across all samples are summarized in Tab. \ref{tab:onlineresults}. On average, our framework eliminates 313.8 out of 704 verification steps, reducing verification by approximately 44.57\% and saving a total of 1569 seconds. Despite this reduction, classification precision remains high. In particular, with $l$-Value = 100\%, false positives decrease by 90.32\% compared to $l$-Value = 50\%, demonstrating a substantial improvement in classification precision while minimizing unnecessary verification steps.
The overall classification results across all samples are summarized in Tab. \ref{tab:onlineresults}. On average, our framework eliminates 302.3 out of 704 verification steps, reducing verification by approximately 42.94\% and saving a total of 1511 seconds.
As such, of the 8 hours 48 minutes needed to record all this data, we could have actually saved around 25 minutes.

Despite this reduction, classification precision remains high. In particular, with $l$-Value = 100\%, false positives instances decrease by 94.44\% compared to $l$-Value = 50\%, demonstrating a substantial improvement in classification precision while minimizing unnecessary verification steps.

\subsection{Post-hoc analysis}
We performed a small grid search for different parameters in the $k$-NN context with $l$-Value = 100\% and we obtained the mean results of 30 runs in the prototype online environment in Tab. \ref{tab:gridsearchresults}. We notice that the best distance metric is the cosine with respect to both precision and the low number of false positives. By using $k=11$, we can balance between the number of false positives and the number of uncertain instances. This small grid search confirms our parameters choices in Sec. \ref{subsec:votingrule}.

\begin{table}[t]
    \vspace{2.4mm}

    \addtolength{\abovecaptionskip}{1.8mm}
    \caption{Results of Precision, True Positive, False Positive, True Negative, False Negative and Uncertain samples of the simulated online self-supervised learning framework with different number of $k$ nearest neighbors and metric distance.}
    \centering
    % \small
    \resizebox{\columnwidth}{!}{
    \begin{tabular}{|c|c|c|c|c|c|c|c|c}
    \hline
    K & Metric & Prec.\% & TP & FP & TN & FN & Unc.\\
    \hline
    \hline
    11 & Cosine & 97.72\% & 110.2 & 2.6 & 182.8 & 6.4 & 401.7\\
    11 & Euclidean & 91.57\% & 57.9 & 5.2 & 144.6 & 4.1 & 491.9\\
    11 & Manhattan & 83.78\% & 51.6 & 11.0 & 60.2 & 2.9 & 578.1\\
    11 & Minkowski & 92.14\% & 54.1 & 4.5 & 152.0 & 4.5 & 488.8\\
    \hline
    5 & Cosine & 95.77\% & 155.4 & 6.9 & 257.5 & 18.6 & 265.5\\
    11 & Cosine & 97.72\% & 110.2 & 2.6 & 182.8 & 6.4 & 401.7\\
    15 & Cosine & 97.26\% & 86.6 & 2.5 & 146.3 & 3.3 & 465.1\\
    21 & Cosine & 98.23\% & 60.2 & 1.1 & 106.7 & 2.4 & 533.5\\
    25 & Cosine & 97.48\% & 59.0 & 1.56 & 113.8 & 2.3 & 527.2\\
    \hline
    \end{tabular}
    }
\label{tab:gridsearchresults}
\end{table}

\section{CONCLUSION}
\label{sec:conclusion}

%%%%%%%%%%%%%%%%%%%%%%%%%%%%%%%%%%%%%%%%%%%%%%%%%%%%%%%%%%%%%%%%%%%%%%%%%%%%%%%%
\addtolength{\textheight}{-8cm}   % This command serves to balance the column lengths on the last page of the document manually. It shortens the textheight of the last page by a suitable amount. This command does not take effect until the next page so it should come on the page before the last. Make sure that you do not shorten the textheight too much.
%%%%%%%%%%%%%%%%%%%%%%%%%%%%%%%%%%%%%%%%%%%%%%%%%%%%%%%%%%%%%%%%%%%%%%%%%%%%%%%%

This study serves as an initial investigation into the feasibility of using simple machine learning models to improve classification precision in industrial applications. As demonstrated in Fig. \ref{fig:resultsTraining}, our framework reduces overall cycle time and simultaneously has high precision. These findings highlight the potential of our approach for industrial applications without compromising performance.

% \subsection{\textcolor{red}{Remarks on Results}}
% Our experimental results demonstrate that integrating force sensing with self-supervised learning significantly enhances verification precision, reducing false positives and minimizing the usage of costly verification methods in the pipeline workflow.

% This approach offers several advantages, including fully automated labeling without human intervention and minimal setup requirements, enabling rapid implementation and deployment. Moreover, the system continuously improves its classification performance as it encounters new data instances, progressively reducing reliance on the time-intensive height verification process Fig. \ref{fig:pipeline}(d).

Our experimental results demonstrate that integrating force sensing with self-supervised learning significantly reduces the usage of costly verification methods in the pipeline workflow as seen in Fig. \ref{fig:pipeline}(e).

This approach offers several advantages, including fully automated labeling without human intervention and minimal setup requirements, enabling rapid implementation and deployment, and the system continuously improves its performance as it encounters new instances.

Tab. \ref{tab:onlineresults} shows that,
% BEFORE
% our method substantially reduces false positives that could otherwise \textcolor{red}{lead to malfunctioning medical injection devices}. 
while false positive instances are not entirely eliminated, it is considerably low. Out of a total of 704 samples, only 2.6 were classified as false positives, representing just 0.36\% of the full dataset. This outcome is notable, especially considering that several optimization challenges remain unaddressed.

For its simplicity and easy understanding, we opted to use $k$-NN instead of more complex models. The $k$-NN algorithm provides greater control over the $l$-value, making it a practical choice for our framework. However, it would be valuable to compare its performance against deep learning models, particularly in relation to dataset size and generalization capabilities.

% \subsection{Future works}
Several challenges remain open for future research. First, the selection of the $k$ parameter in $k$-NN was not rigorously optimized as we used a fixed range of values for the optimal configuration. Second, ensuring the quality of the data added to the growing dataset remains an issue. Although the height verification process provides additional labeled data, not all instances should be incorporated, but only those that enhance classification precision should be retained. 
% Addressing these issues is crucial for improving long-term system performance.

The dataset presented in Fig. \ref{fig:Dataset} reveals significant variations and the presence of outliers in both positive and negative samples. Collecting more data, particularly for rare or outlier cases, could improve model generalization and overall classification performance. Additionally, our pre-processing pipeline (Sec. \ref{subsec:datatcollection}) may have inadvertently removed important features during the sliding window down-sampling process, or the feature vector may still be too large, impacting efficiency.

To further improve the framework, we plan to:

\begin{itemize}
    \item Gather more force data of the insertion task.
    \item Investigate alternative deep learning models to identify the most effective architecture for our case of force-based classification.
    \item Refine data selection strategies to construct a more efficient and compact dataset, reducing computational overhead while maintaining classification precision.
    \item Extend the framework to different test objects to evaluate its generalizability.
\end{itemize}

% Anopther important modifications is to put a threshold in the distance metric of the k-NN as well. It might be possible that the FP happens mostly because the nearest neighbours of the new test data are very far away, meaning that the class assignment is not that precise. Distance is an important metric.
%%%%%%%%%%%%%%%%%%%%%%%%%%%%%%%%%%%%%%%%%%%%%%%%%%%%%%%%%%%%%%%%%%%%%%%%%%%%%%

\section*{ACKNOWLEDGMENT}

This project was funded in part by Innovation Fund Denmark through the project FERA (3149-00014A), and in part by the SDU I4.0-Lab.

%%%%%%%%%%%%%%%%%%%%%%%%%%%%%%%%%%%%%%%%%%%%%%%%%%%%%%%%%%%%%%%%%%%%%%%%%%%%%%%%
% \IEEEtriggeratref{20}
% \enlargethispage{+20mm}
\bibliographystyle{IEEEtran}
\bibliography{References}
\clearpage
\end{document}